\documentclass{article}

\PassOptionsToPackage{comma,square,sort&compress,numbers}{natbib}

\usepackage[preprint]{neurips_2025}

\usepackage[utf8]{inputenc} 
\usepackage[T1]{fontenc}    
\usepackage[dvipsnames, x11names]{xcolor}        
\definecolor{blue}{rgb}{0.21,0.49,0.74}
\usepackage[pagebackref,breaklinks,colorlinks,allcolors=blue]{hyperref}
\usepackage{url}            
\usepackage{booktabs}       
\usepackage{amsfonts}       
\usepackage{nicefrac}       
\usepackage{microtype}      

\usepackage{graphicx}
\usepackage{amsmath}
\usepackage{multirow}
\usepackage{enumitem}

\usepackage{orcidlink}

\usepackage{colortbl}

\usepackage{wrapfig}
\usepackage{float}
\usepackage{amssymb}
\usepackage{marvosym}

\usepackage{xspace}
\usepackage{wrapfig}

\newcommand{\ie}{{i.e.}}
\newcommand{\eg}{{e.g.}}

\newcommand{\etal}{\textit{et al.}}

\let\titleold\title
\renewcommand{\title}[1]{\titleold{#1}\newcommand{\thetitle}{#1}}
\def\maketitlesupplementary
{
    \begin{center}
        \Large
        \textbf{\thetitle}\\
        \vspace{0.5em}Supplementary Material \\
        \vspace{1.0em}
    \end{center}
}

\title{Cross from Left to Right Brain: Adaptive Text Dreamer for Vision-and-Language Navigation}

\author{
  \hspace{-9pt}Pingrui Zhang$^{1, 2}$, \hspace{-1pt}
  Yifei Su$^{3, 4}$, \hspace{-1pt}
  Pengyuan Wu$^2$, \hspace{-1pt}
  Dong An$^5$, \hspace{-1pt}
  Li Zhang$^6$, \hspace{-1pt}
  \textbf{Zhigang Wang}$^2$, \hspace{-1pt}\\
  \textbf{Dong Wang}$^2$, \hspace{-1pt}
  \textbf{Yan Ding}$^2$, \hspace{-1pt}
  \textbf{Bin Zhao}$^2$, \hspace{-1pt}
  \textbf{Xuelong Li}$^7$$^\dag$\\
  \hspace{-0.4cm}$^1$Fudan University \hspace{0pt}
  $^2$Shanghai AI Laboratory \\
  $^3$School of Artificial Intelligence, University of Chinese Academy of Sciences \hspace{0pt}\\
  $^4$MAIS, Institute of Automation of Chinese Academy of Sciences \hspace{0pt}\\
  $^5$Mohamed bin Zayed University of Artificial Intelligence \hspace{0pt}\\
  $^6$University of Science and Technology of China \hspace{0pt}\\
  $^7$TeleAI, China Telecom Corp Ltd \hspace{0pt}\\
  \tt\small{zhangpingrui@pjlab.org.cn, xuelong\_li@ieee.org}
  \hspace{-0.4cm}
}

\begin{document}

\maketitle

\begin{abstract}

Vision-and-Language Navigation (VLN) requires the agent to navigate by following natural instructions under partial observability, making it difficult to align perception with language. 
Recent methods mitigate this by imagining future scenes, yet they rely on vision-based synthesis, leading to high computational cost and redundant details.
To this end, we propose to adaptively imagine key environmental semantics via \textit{language} form, enabling a more reliable and efficient strategy. Specifically, we introduce a novel Adaptive Text Dreamer (ATD), a dual-branch self-guided imagination policy built upon a large language model (LLM). 
ATD is designed with a human-like left-right brain architecture, where the left brain focuses on logical integration, and the right brain is responsible for imaginative prediction of future scenes. 
To achieve this, we fine-tune only the Q-former within both brains to efficiently activate domain-specific knowledge in the LLM, enabling dynamic updates of logical reasoning and imagination during navigation.
Furthermore, we introduce a cross-interaction mechanism to regularize the imagined outputs and inject them into a navigation expert module, allowing ATD to jointly exploit both the reasoning capacity of the LLM and the expertise of the navigation model.
We conduct extensive experiments on the R2R benchmark, where ATD achieves state-of-the-art performance with fewer parameters.
The code is \href{https://github.com/zhangpingrui/Adaptive-Text-Dreamer}{here}.

\end{abstract}

\section{Introduction}
\label{sec:intro}

\begin{figure*}[!tbp]
\centering
\includegraphics[width=0.99\linewidth]{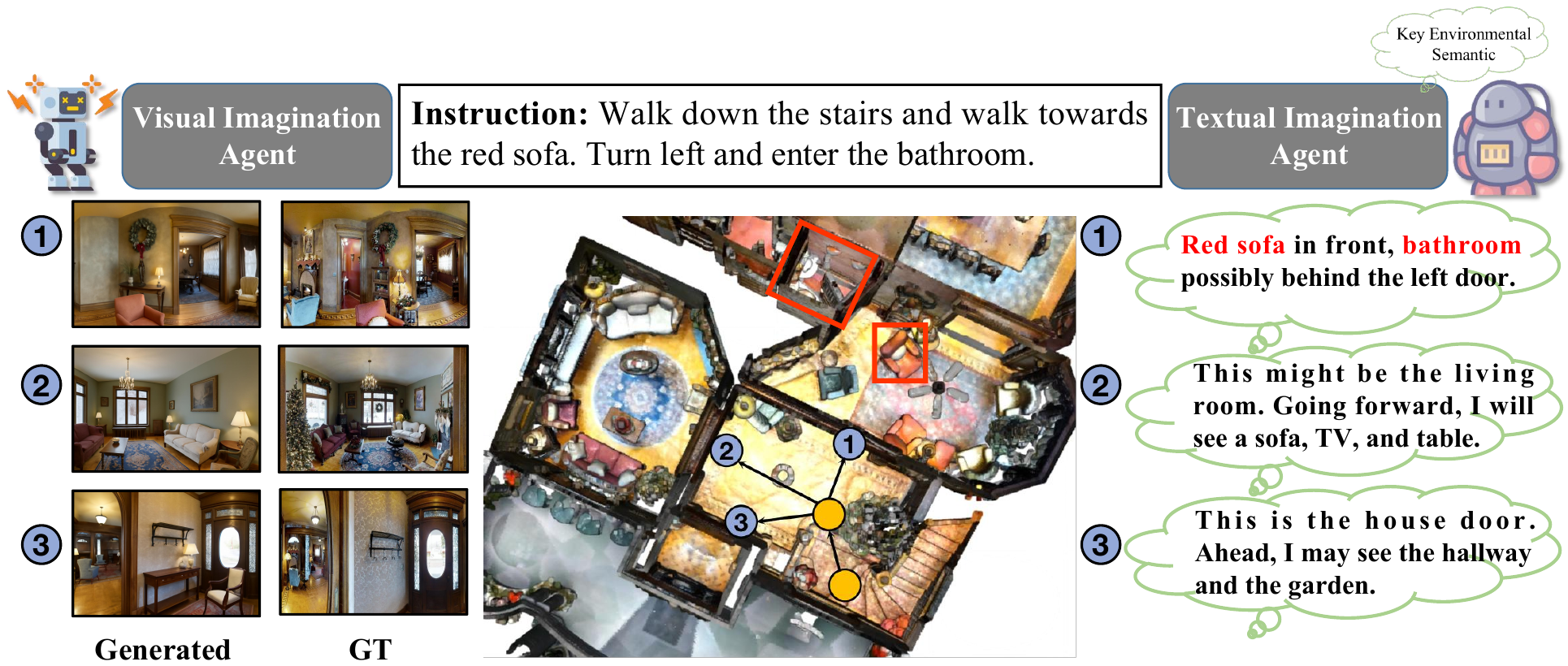}
\caption{Given the task ``Walk down the stairs and walk towards the red sofa. Turn left and enter the bathroom'', and assuming $3$ candidate viewpoints are encountered during the current navigation step. Left: Existing methods generate the visual imagination for each candidate. While effective, the rendered images often contain blurry from GT or redundant regions, increasing both imagination time and alignment difficulty. Right: Our ATD predicts the key future semantics (\textcolor{red}{red color}) in textual form, thereby facilitating the construction of compact yet semantically rich future representations with improved efficiency. \includegraphics[scale=0.025]{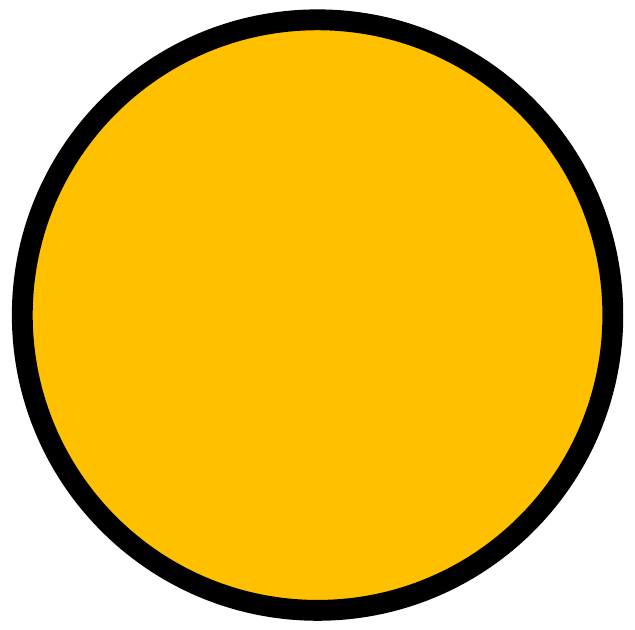}: Visited Node. \includegraphics[scale=0.025]{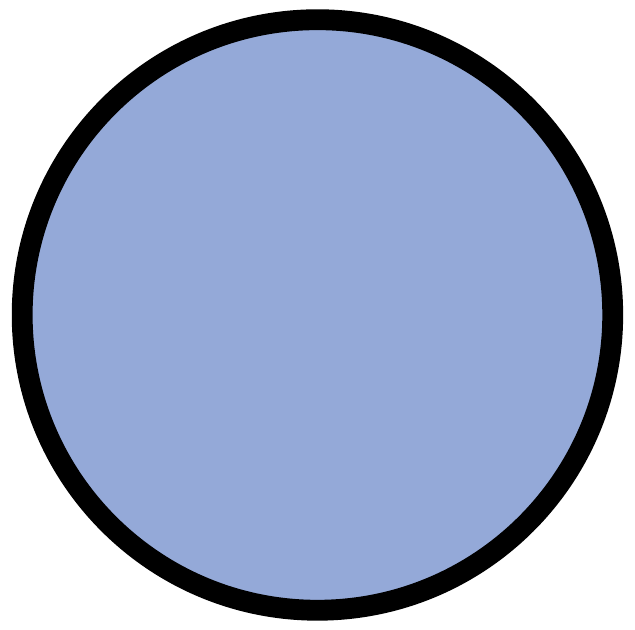}: Candidate Node. \includegraphics[scale=0.025]{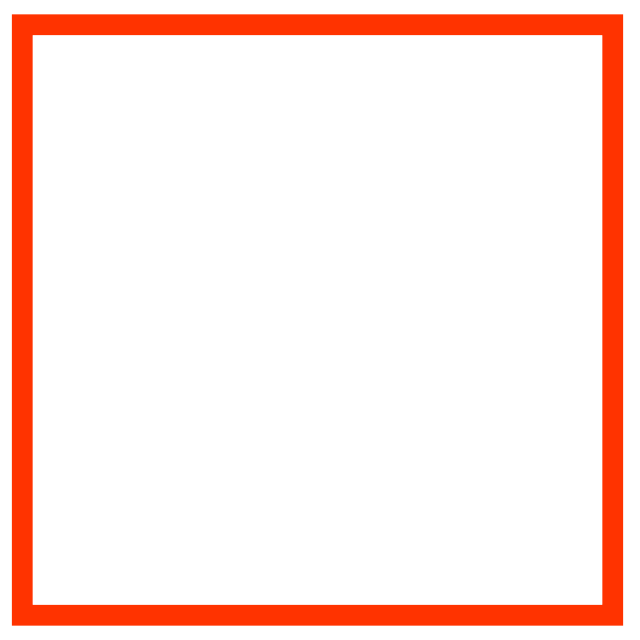}: Key Environmental Semantic.}
\label{fig:intro}
\vspace{-6mm}
\end{figure*}

Vision-and-Language Navigation (VLN)~\cite{r2r} is a popular Embodied AI task where an agent follows natural language instructions to reach a goal in previously unseen 3D environments.
A fundamental challenge of VLN lies in its partially observable nature—agents perceive only a limited field of view at each step, while relevant landmarks or goals may lie far beyond.
This restricted observability introduces substantial uncertainty in grounding language to the environment, requiring the agent to reason beyond immediate perception and infer unobserved semantic cues for successful navigation.

To address this uncertainty, early approaches introduced memory-based mechanisms~\cite{rcm,hamt,duet,recurrent_vln,gridmm} to help agents aggregate past observations or build spatial representations of traversed regions.
While beneficial, memory alone is insufficient for anticipating critical, yet-unseen information, especially when reasoning about instruction-relevant targets outside the current or historical view.
To overcome this, recent works have proposed imagination-based strategies~\cite{kerm,foresightnav} to extend the agent’s perceptual horizon by simulating observations of unvisited locations. These approaches enable agents to proactively infer what might lie ahead, reducing uncertainty and improving long-horizon planning.

While promising, existing imagination-based methods primarily rely on visually grounded generation, which presents several limitations. 
For example, DREAMWALKER~\cite{dreamwalker} generates pixel-level images along candidate trajectories, incurring high computational cost and suffering from noise accumulation.
HNR~\cite{hnr} improves efficiency through feature-level synthesis, yet still depends on encoding full panoramic views, often introducing redundant or distracting information. In contrast, we argue that effective navigation hinges not on full-scene reconstruction, but on identifying task-relevant semantics. 
For example, given the instruction ``Walk down the stairs and walk towards the red sofa.'', the core challenge lies in inferring where the \texttt{red sofa} might appear—while other visual content remains largely irrelevant (Fig.~\ref{fig:intro}). This insight motivates a shift toward more abstract, selective imagination.
Language, with its compositional and abstract representation, is naturally suited to support such targeted imagination~\cite{lin2023learning}.

Motivated by this, we propose to imagine the environment through language and introduce \textbf{A}daptive \textbf{T}ext \textbf{D}reamer (\textbf{ATD}), a dual-branch LLM self-guided dynamic imagination policy for VLN. The model consists of two branches of LLMs: one estimates the current navigation state (\textbf{Left brain: State Estimation LLM}), while the other generates imaginative descriptions for the future pathway (\textbf{Right brain: Imagination LLM}). Similar to the human brain, the left hemisphere is responsible for logical integration, while the right hemisphere excels in divergent thinking. For both the left and right brains, we use Q-Former~\cite{blip} to perform visual instruction tuning on the frozen InstructBLIP~\cite{instructblip}, using the respective text-predicted ground truth collected for each brain's function, thereby activating the corresponding domain's commonsense knowledge within the LLM. During navigation, the left brain constrains the imagination of the right brain. This architecture allows the imagination of the text dreamer to be continuously updated throughout the navigation process, preventing overemphasis on the completed parts of the navigation instruction, which could otherwise negatively impact the imagination. To bridge the LLM with the policy, we use the latent adaptive cross-interactive imagination embedding of ATD to inject into the node embedding of the action expert policy. This latent alignment structure allows us to preserve the rich linguistic reasoning capabilities of the LLM while fully leveraging the specialized skills of the navigation action expert~\cite{forewarn, pi_0, bu2025agibot}. 

We evaluate our proposed ATD on the R2R~\cite{r2r} benchmark and observe clear improvements over prior approaches with fewer parameters. 
Specifically, our method achieves gains in Success Rate (\texttt{SR}) of 8.0\% and 12.0\%, and in Success weighted by Path Length (\texttt{SPL}) of 5.0\% and 11.0\%, on the \texttt{val seen} and \texttt{val unseen}, respectively. 
These results highlight the efficiency and effectiveness of our proposed \textbf{Adaptive Text Dreamer} guided navigation.
Our contributions are summarized as follows: 
(1) We highlight the advantage of using language for future imagination and propose a human-like left-right brain structure to enable an adaptive text dreamer. This approach utilizes one LLM to guide another, resulting in enhanced and more accurate imaginative capabilities. 
(2) We design a scheme to train the State Estimation LLM and Imagination LLM to predict the key environmental semantics during navigation.
(3) Our method achieves state-of-the-art performance on both \texttt{SR} and \texttt{SPL} metrics, while maintaining fewer parameters.

\section{Related work}
\label{sec:related}

\paragraph{Vision-and-language Navigation (VLN).}\label{sec:related_vln}
VLN requires agents to navigate complex environments via visual clues to follow various human instructions, e.g., step-by-step commands~\cite{r2r,rxr,touchdown}, high-level instructions~\cite{reverie,citynav,alfred}, and chat dialog~\cite{cvdn,avdn,hanna,r2h}. Impressive progress has been achieved mainly by three factors: 
(1) Data Augmentation. 
Some work~\cite{envedit, panogen, panogen++} benefit from augmenting the scenes and observations, while others~\cite{unseenfromseen, bevinstructor} resort to enriching the instructions. Besides, the speaker-follower paradigm~\cite{sf, envdrop, ccc, scale_vln, lana} effectively scales VLN performance via abundant instruction-path pairs.
(2) Planning Strategy. 
Beyond classical imitation learning~\cite{r2r, newpath, dagger, duet} and reinforcement learning~\cite{rcm, recurrent_vln, softexpert} strategies, various auxiliary tasks~\cite{selfmonitoring,auxvln,regretful_vln}, explicit external knowledge~\cite{kerm,ckr,marchinchat}, and rollback mechanisms~\cite{tactical,regretful_vln} further advance the VLN performance.
(3) Representation Alignment Learning. 
Well-designed historical observation representation, \eg, recurrent vectors~\cite{recurrent_vln}, trajectory views~\cite{hamt,et}, topological map~\cite{duet,egp}, grid maps~\cite{bevbert, gridmm}, and feature volume~\cite{ver_vln}, offer solid foundation for representation alignment in VLN. Building on this, various pertaining based on proxy-tasks~\cite{prevalent,hop,hop+,duet,bevbert,planningfromimg,zhang2025moma} and vision-language models~\cite{navillm,leo} have significantly boosted VLN performance.

\paragraph{Imagination in VLN.}\label{sec:related_imagine}
Given the partially observable Markov property of VLN, recent works introduce future scene imagination to enhance multimodal alignment~\cite{dreamwalker,hnr,unitedvln,nie2025wmnav}. Initially, MIND~\cite{mind} generates future sub-goal images prior to decision-making to facilitate embodied QA~\cite{eqa,zhang2024question}. PathDreamer~\cite{pathdreamer} built a world model to predict future trajectory images and integrate a text-path compatibility model~\cite{compatibility} for VLN. Similarly, VLN-SIG~\cite{sig_vln} and ImagineNav~\cite{imaginenav} adopt visual cookbook techniques and generative models, respectively, to imagine future candidate observations and make decisions via similarity measures or LLM reasoning. DreamWalker~\cite{dreamwalker} further extends this paradigm to continuous environments~\cite{vlnce} and leverages Monte Carlo Tree search~\cite{mcts1,mcts2} for decision-making. NWM~\cite{nwn} first scales the navigation world model to the 1B level, enabling long-horizon imagination in complex real environments. HNR~\cite{hnr} and UnitedVLN~\cite{unitedvln} leverage the advanced volumetric rendering~\cite{nerf,mipnerf,instantngp} and 3DGS~\cite{3dgs,pixelsplat,mvsplat} to generate future observations, respectively, boosting VLN-CE performance. Beyond predicting RGB, some works assist navigation by imagining semantic maps~\cite{sgm}, geometric maps~\cite{foresightnav}, and hybrid imagination-reality memory~\cite{planningfromimg} of scenes. This work avoids costly future generations by encouraging the model’s right brain to imagine candidate node semantics during training, implicitly encoding future information without accessing future poses.

\paragraph{VLN with LLM.}\label{sec:related_vln_llm}
LM-Nav~\cite{lmnav} uses GPT-3~\cite{gpt3} to extract landmarks from instructions to assist navigation. MiCs generate detailed plans from static and dynamic perspectives as knowledge. LLM-Planner~\cite{llm_planner} and SayNav~\cite{saynav} utilize LLMs~\cite{gpt3,achiam2023gpt} as core planners to dynamically plan high-level actions. As for VLN, NavGPT~\cite{navgpt} builds an LLM agent that converts tasks, observations, and actions into text for zero-shot decision-making. Subsequent works further integrate this paradigm with map-guided prompting~\cite{mapgpt}, multi-expert discussions~\cite{discussnav}, memory maps~\cite{mcgpt}, and hierarchical reasoning~\cite{flexvln}. Other works explore applying LLMs in continuous environments. A$^2$Nav~\cite{a2nav} uses GPT-3~\cite{gpt3} to decompose instructions into a few sub-tasks and trains separate navigators for execution. InstructNav~\cite{instructnav} and CA-Nav~\cite{canav} extend the decomposition paradigm with versatile value maps to guide navigation. AO-Planner~\cite{aoplanner} and Select2Plan~\cite{select2plan} design navigable points within images as prompts to guide VLM to navigate. 
Benefiting from the continuous development of parameter-efficient fine-tuning techniques~\cite{houlsby2019parameter,hu2022lora,jia2022visual,wang2024mos}, some work fine-tune VLMs with in-domain VLN data for better performance. Zheng \etal~\cite{navillm} trains a general navigator via casting various tasks into a unified QA form. Subsequent efforts~\cite{navid,uninavid,navila} extend it to continuous environments by predicting meta-actions~\cite{vlnce} or waypoints~\cite{waypointce} in text form. MapNav~\cite{mapnav} and P3Nav~\cite{p3nav} further explore the utility of semantic maps and adaptive historical sampling as inputs. Beyond predicting actions, introducing intermediate reasoning tasks during navigation ~\cite{flame,navgpt2} can enhance interpretability.

\section{Method}\label{sec:method}

\begin{figure*}[t]
\centering
\includegraphics[width=0.99\linewidth]{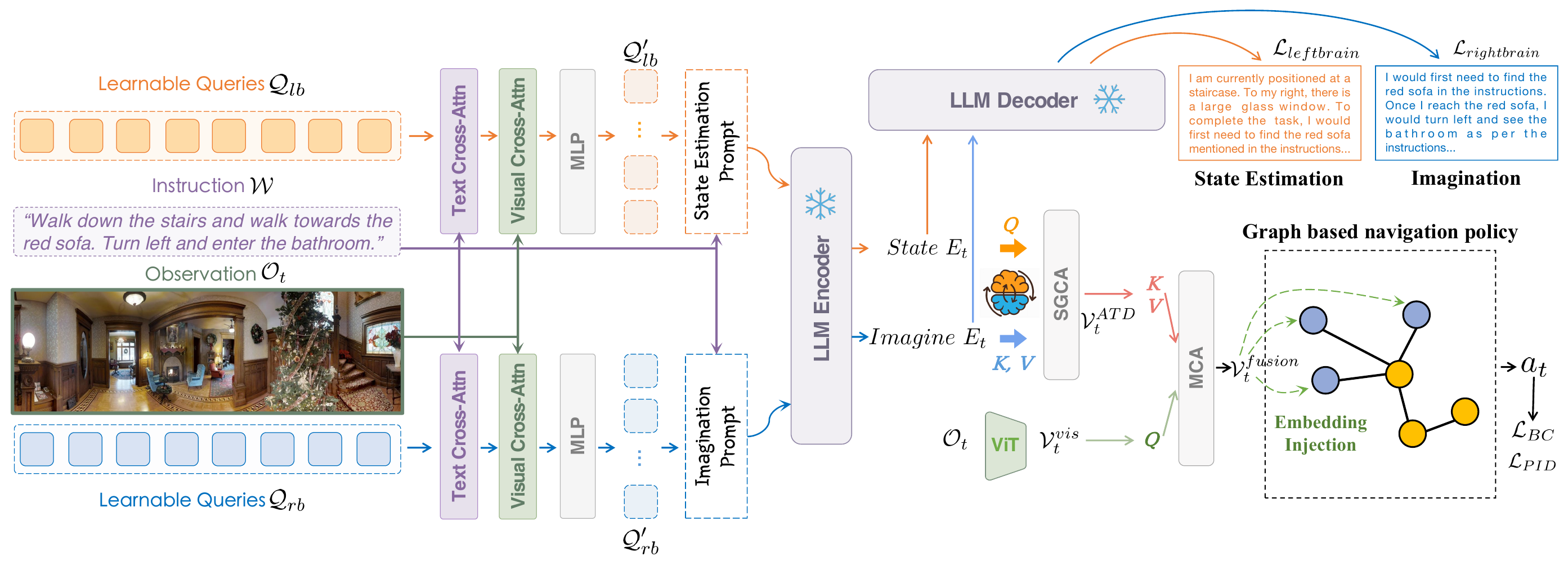}
\vspace{-10pt}
\caption{\textbf{Overview of the Adaptive Text Dreamer (ATD) architecture.} The dual-branch left–right vision-language model (VLM) structure consists of a left-brain branch that performs state estimation using a large language model (LLM), and a right-brain branch that generates future candidate imaginations constrained by latent embeddings derived from the state estimation. The imagined latent representations are then integrated into a graph-based navigation expert to guide action decisions. }
\label{fig:method}
\end{figure*}

As illustrated in Fig.~\ref{fig:method}, the proposed Adaptive Text Dreamer (ATD) architecture consists of a dual-branch left–right reasoning structure and a graph-based navigation expert. We design the left-right VLM branches to support the VLN task in a complementary manner. The left-brain VLM employs a Q-Former module to extract image tokens, which are then fed into a large language model (LLM) to generate state estimation (Sec.\ref{subsec:separate_llm}). The right-brain VLM shares the same architecture as the left brain but is responsible for generating imagination of future candidates (Sec.\ref{subsec:separate_llm}). To ensure the imagination within the current context, we constrain it via latent embeddings derived from the state estimation. The attended imagination latent is then injected into the graph-based navigation policy to guide the action decision process (Sec.~\ref{subsec:ATD}).

\paragraph{Task Setup.} 
In Vision-and-Language Navigation (VLN), the environment is represented as an undirected graph $\mathcal{G} = \{\mathcal{V}, \mathcal{E}\}$, where $\mathcal{V}=\{V_i\}_{i=1}^{K}$ denotes the set of $K$ navigable nodes, and $\mathcal{E}$ represents the connectivity edges between nodes. The agent is initialized at a starting node $V_0 \in \mathcal{V}$ and is provided with a natural language instruction $\mathcal{W} = \{w_i\}^L_{i=1}$ consisting of $L$ words. The goal of the agent is to navigate through the graph based on the instruction and reach the target location or object specified by the instruction. At each time step $t$, the agent perceives its surrounding environment by observing a set of RGB views $\mathcal{O}_t \triangleq {\langle o_i, a_i \rangle}_{i=1}^N$ for each connected navigable node candidate. Here, $N$ denotes the number of candidate nodes, where each view $o_i$ (for $i \leq N$) is associated with a direction angle $a_i$ relative to the agent's heading. The agent selects the next action by predicting the relative angle $a_t$ from $\mathcal{O}_t$. The agent's policy $\pi$, parameterized by ${\Theta}$, is learned to predict the action $a_t$ based on the given instruction $\mathcal{W}$ and the observed set of views $\mathcal{O}_t$, \ie, $\pi(a_t|\mathcal{W}, \mathcal{O}_t; {\Theta})$.

\subsection{State Estimation LLM and Imagination LLM}
\label{subsec:separate_llm}
In this section, we introduce the process of collecting instruction data and fine-tuning frozen InstructBLIP~\cite{instructblip} with Q-former~\cite{blip} to obtain the two ``brains'' of our ATD model. The training data for these two LLMs is collected in the discrete simulation of VLN, as it has a predefined graph that allows us to pre-extract visual features for captioning or reasoning.

\paragraph{State Estimation LLM.} State estimation is always an important and challenging problem in VLN, as determining the current stage of navigation within the instruction helps to exclude interference from completed parts of the text information. Some previous work~\cite{lin2024correctable} uses object detection to update the landmark list as a representation of state updates. However, this approach is rigid and lacks flexibility, as it does not involve true reasoning to assess the progress of navigation. Simply determining whether an object appears in the field of view may overlook whether the object actually matches the spatial relationship specified for the landmark in the instruction, thus failing to ensure that it is indeed the object described in the instruction. 

Thus, the flexible reasoning ability of the LLM is leveraged for state estimation. Following ~\cite{navgpt2}, the state description is collected by feeding observations and instructions into GPT-4V~\cite{achiam2023gpt}. In the prompt, GPT-4V is tasked with reasoning about the agent's current position in the environment based on a single visual input and the navigation instructions, while determining the current state of the navigation. The detailed prompt can be found in the appendix. The reasoning output $\mathcal{R}_{t}$ of GPT-4V is then used to fine-tune an LLM that with fewer parameters, distilling the state estimation capability into the model. During the training process of state estimation, the model learns to predict the current state based on the observations and instructions, i.e., the left brain. Specifically, the process of training the State Estimation LLM, including its inputs and outputs, can be formulated as follows: 
\begin{equation}
    \mathcal{Q}'_{lb} = \mathrm{Q}\mbox{-}\mathrm{former}^{lb}( \mathcal{W}, \mathcal{O}_t, \mathcal{Q}_{lb}), 
\label{eq:leftbrain qformer}
\end{equation}
\begin{equation}
    \langle \hat{\mathcal{R}}_{t}, State\ E_{t} \rangle = \mathrm{LLM}_{frozen}( \mathrm{StateEsimationPrompt}( \mathcal{W}, \mathcal{Q}'_{lb}) ). 
\label{eq:state estimation llm}
\end{equation}

\paragraph{Imagination LLM.} As mentioned below in Sec.~\ref{sec:intro}, in order to represent the prediction of future key environmental semantics using highly abstract language, it is essential to train an LLM as a text dreamer, i.e., the right brain. To achieve this, we formulate the process of generating imaginative representations as follows: 
\begin{equation}
    \mathcal{Q}'_{rb} = \mathrm{Q}\mbox{-}\mathrm{former}^{rb}( \mathcal{W}, \mathcal{O}_t, \mathcal{Q}_{rb}), 
\label{eq:rightbrain qformer}
\end{equation}
\begin{equation}
    \langle \mathcal{I}_{t}, Imagine\ E_{t} \rangle = \mathrm{LLM}_{frozen}( \mathrm{ImaginationPrompt}( \mathcal{W}, \mathcal{Q}'_{rb}) ), 
\label{eq:imagination llm}
\end{equation}
where $\mathcal{Q}_{lb}, \mathcal{Q}_{rb} \in \mathbb{R}^{n \times d_1}$ represents the learnable query tokens for $\mathrm{Q}\mbox{-}\mathrm{former}$, $\mathcal{Q}'_{lb}, \mathcal{Q}'_{rb} \in \mathbb{R}^{n \times d_1}$ denotes the text-visual fusion output from the left-right brain branch, $\hat{\mathcal{R}_t}, \mathcal{I}_{t}$ is the reasoning and imagination output generated by the LLM at time step $t$, and $State\ E_{t}, Imagine\ E_{t} \in \mathbb{R}^{m \times d_2}$ represents the hidden state and imagine embedding feature of the LLM. Both $\mathrm{StateEstimationPrompt}$ and $\mathrm{ImaginationPrompt}$ are designed to integrate visual and textual information for input to subsequent LLMs. The specific details of the prompts can be found in the appendix.

To obtain the Imagination LLM, long and detailed captions $\{\mathcal{C}^i_{candidate\_t}\}_{i=1}^{N}$ for $N$ candidate nodes at each sampled location in the trajectory are collected using Qwen2.5-VL-7B~\cite{bai2025qwen2} as the ground truth for the text dreamer. Then, the ImaginationPrompt directs the LLM to imagine what scenes or objects might appear in different directions based on the current node's observation and the given instruction for the current episode.

\paragraph{Loss.} The loss is computed at each sampled point in the trajectory to train the two branches of the LLM. Cross-entropy loss is used during training of the brains, as shown in the following equations: 
\begin{equation}
    \mathcal{L}_{leftbrain} = - \sum_{t=1}^{T} \mathcal{R}_{t} \log(\hat{\mathcal{R}}_{t}),\quad \mathcal{L}_{rightbrain} = - \sum_{t=1}^{T} \sum_{i=1}^{N}\{\mathcal{C}^i_{candidate\_t}\} \log(\mathcal{I}_{t}), 
\label{eq:cross-entropy loss}
\end{equation}
After this training stage, we obtain the left and right brain LLMs, which are used to guide the subsequent navigation policy and further enhance the model's decision-making.

\subsection{Adaptive Text Dreamer}
\label{subsec:ATD}
After the aforementioned training, the State Estimation LLM and Imagination LLM are obtained. In this section, these two LLMs are integrated as the two brains into the navigation policy for training, resulting in the final navigation policy—Adaptive Text Dreamer (ATD). 

\paragraph{State Grounded Imagination.} The imagination of the right brain is unconstrained, allowing it full freedom to envision future scenarios. At the same time, to prevent irrelevant details from the imagination from excessively interfering with navigation, the imagination is constrained using the navigation state estimation information. Thus, state information is employed to ground the important and relevant details from the imagination through State Grounded Cross-Attention (SGCA) layers. We serve state estimation embedding $State\ E_{t}$ as queries and the imagination embedding $Imagine\ E_{t}$ as keys and values, and then compute their cross-attention weight matrix $\mathbf{A} \in \mathbb{R}^{n \times N\times M}$~\cite{vaswani2017attention}. The process is formulated as follows:
\begin{equation}
    \mathbf{Q}_{S} = State\ E_{t}\mathbf{W}^{Q}, \mathbf{K}_{I} = Imagine\ E_{t}\mathbf{W}^{K}, \mathbf{V}_{I} = Imagine\ E_{t}\mathbf{W}^{V}
\label{eq:qkv}
\end{equation}
\begin{equation}
    \mathbf{A} = \operatorname{SoftMax}(\operatorname{Sim}_{\cos}(\mathbf{Q}_{S}, \mathbf{K}_{I})),
\end{equation}
where $\mathbf{Q}_S \in \mathbb{R}^{N_q \times d}, \mathbf{K}_{I}, \mathbf{V}_{I} \in \mathbb{R}^{n \times M_{kv}\times d}$. $\operatorname{Sim}_{\cos}$ denotes the cosine similarity function. The attention matrix $\mathbf{A} \in \mathbb{R}^{n \times N_q\times M_{kv}}$ represents the constraint weights from state estimation to imagination, guiding the refinement of the information in the imagination. Thus, when the state changes, $\mathbf{A}$ adjusts accordingly to filter the imagination information. This adaptive refinement process, known as SGCA, can be mathematically expressed as: 
\begin{equation}
    \operatorname{SGCA}(\mathbf{Q}_S,\mathbf{K}_{I},\mathbf{V}_{I}) = \mathbf{A}\cdot \mathbf{V}_{I}.
\label{eq:sgca}
\end{equation}
The grounded feature is then fed into the node embedding of the subsequent navigation policy.

\paragraph{Graph-based Navigation Policy.} To ensure efficient action planning, a navigation policy based on a dynamic topological graph~\cite{duet} is employed, which updates in real-time to encode navigation history through nodes and edges. Edges record traversable paths, allowing the graph to explicitly represent the environmental topology and spatial relationships. Nodes are categorized into visited locations, navigable neighbors, and the current position, with new nodes (e.g., unvisited adjacent states) added incrementally as the agent explores. The policy utilizes a multi-layer transformer to update each node's visual embeddings, denoted as $\mathcal{V}^{vis}_{t}$, which represents the average pooling of the observed view representations surrounding each node.

\paragraph{Latent Embedding Injection.} The output of SGCA is the ATD embedding $\mathcal{V}^{ATD}_{t}$ of the node. The adaptive imagination information is then integrated into the node embedding of the policy graph. Multi-head Cross-Attention (MCA) is used to fuse these two node embeddings, with $\mathcal{V}^{vis}{t}$ as the query and $\mathcal{V}^{ATD}_{t}$ as both the key and value. This process is formulated as follows: 
\begin{equation}
    \mathcal{V}^{fusion}_{t} = \text{MCA}(\mathcal{V}^{ATD}_{t}, \mathcal{V}^{vis}_{t}),
\label{eq:grounded}
\end{equation}
where the $\mathcal{V}^{fusion}_{t}$ represents the fused embedding of the graph node. At step $t$, a navigation graph $\mathcal{G}_t = {\mathcal{V}_t, \mathcal{E}_t}$ is constructed, where $\mathcal{G}_t \subset \mathcal{G}$. The node embeddings $\mathcal{V}_t$ are passed through a multi-layer cross-modal transformer to capture the relationship between the instructions and the nodes. Initially, the node embeddings undergo cross-attention with the instructions encoded by the LLM. Afterward, a graph-aware self-attention (GASA) layer is applied, which enhances the contextual understanding by considering both the spatial distances and visual similarities between nodes. The GASA operation is formulated as follows:
\begin{equation}
\text{GASA}(\mathcal{V}) = \text{Softmax}\left( \frac{\mathcal{V}W_q(\mathcal{V}W_k)^T}{\sqrt{d}} + EW_e \right)\mathcal{V}W_v,
\end{equation}
$E$ is the pairwise distance matrix derived from the graph edges $\mathcal{E}_t$. The matrices $W_q$, $W_k$, $W_e$, and $W_v$ are learnable parameters. $\mathcal{V}$ represents the node embedding, which, in this case, is $\mathcal{V}^{fusion}$. The output of GASA is then fed into a Multi-Layer Perceptron (MLP) to predict the next goal action. 

\paragraph{Loss.} In the loss computation process, we follow previous imitation learning methods~\cite{ross2011reduction,duet}, using both Behaviour Cloning (BC) loss and Pseudo Interactive Demonstrator (PID) loss to supervise the model. The loss formulas are as follows: 
\begin{equation}
    \mathcal{L}_{\text{BC}} = -\sum^T_{t=1}\log{\pi(a^*_t|\mathcal{W}, \mathcal{G}_t)},\quad \mathcal{L}_{\text{PID}} = -\sum^T_{t=1}\log{\pi(\tilde{a}^*_t|\mathcal{W}, \tilde{\mathcal{G}}_t)},
\end{equation}
where $a^*_t$ represents the ground truth action, and $\tilde{a}^*_t$ denotes the pseudo label, which is determined by the shortest path to the destination from the partial graph $\tilde{\mathcal{G}}_t$ generated by the agent through on-policy action sampling. This pseudo action closely resembles the action of the optimal policy $\pi$, which selects the next target node based on the shortest distance to the final destination using the environment graph $G$. Finally, The total loss function is given by $\mathcal{L} = \lambda \mathcal{L}_{\text{BC}} + \mathcal{L}_{\text{PID}}$, where $\lambda$ acts as a trade-off factor.

\section{Experiment}\label{sec:experiment}

\subsection{Dataset and Evaluation Metrics} \label{sec:exp_data_metric}
We conduct systematic evaluations of the proposed model on the widely used R2R~\cite{r2r} benchmark in discrete environments. The R2R dataset provides step-by-step navigation instructions, with each instruction averaging $32$ words and covering approximately $6$ navigation steps. We evaluate the performance via a comprehensive suite of standard navigation metrics~\cite{r2r,reverie,vlnce}, including: Trajectory Length (\texttt{TL}), measuring the average path length in meters; Navigation Error (\texttt{NE}), denoting the average distance between the final position and destination; Success Rate (\texttt{SR}) the percentage of trajectories with \texttt{NE} less than $3$ meters; Oracle Success Rate (\texttt{OSR}), the success rate under an ideal stopping policy; Success weighted by Path Length (\texttt{SPL})~\cite{evalvln}, combining the success rate and the trajectory efficiency to comprehensively assess navigation performance.

\subsection{Implementation Details.} \label{sec:exp_details}
The proposed ATD framework is built on InstructBLIP~\cite{instructblip} with Flan-T5-xl as LLM decoder. For training the State Estimation and Imagination LLMs, we only fine-tune the Q-former module and keep the LLM and vision backbone frozen with a batch size of $1$. 
All fine-tuning are conducted on a single GPU. The AdamW optimizer is used with a learning rate of $1e\mbox-5$ and weight decay of $0.05$. For training the ATD navigation policy, we again use a single GPU, but with a batch size of $2$ and the AdamW optimizer, keeping the learning rate at $1e\mbox-5$. More detailed experiment settings can be found in the supplementary material.

\subsection{Quantitative Analysis} \label{sec:exp_quantitative}

\begin{table*}[t]
\centering
\caption{\textbf{Performance comparison on the R2R dataset.} \ours\ ooutperforms all previous methods that utilize LLMs and closes the performance gap among state-of-the-art methods trained at a small scale. $\dag$: Indicates methods that leverage additional visual data beyond MP3D.} 
\vspace{5pt}
\resizebox{\textwidth}{!}{
\normalsize
\definecolor{Gray}{gray}{0.94} 

\begin{tabular}{lccccccccc>{\columncolor{Gray}}c>{\columncolor{Gray}}cccc>{\columncolor{Gray}}c>{\columncolor{Gray}}c} %

\toprule
\midrule
\multicolumn{1}{c}{\multirow{2}{*}{\textbf{Methods}}} & \multicolumn{1}{c}{\multirow{2}{*}{\parbox{1cm}{\vspace*{1ex}Freeze \\ \vspace*{1.5ex} LLM}}} & 
\multicolumn{5}{c}{Val Seen} & 
\multicolumn{5}{c}{Val Unseen} & 
\multicolumn{5}{c}{Test Unseen} \\ 

\cmidrule(r){3-7} 
\cmidrule(r){8-12} 
\cmidrule(r){13-17} 

\multicolumn{2}{c}{} & 
\multicolumn{1}{c}{TL} & 
\multicolumn{1}{c}{NE$\downarrow$} & 
\multicolumn{1}{c}{OSR$\uparrow$} & 
\multicolumn{1}{c}{\textbf{SR}$\uparrow$} & 
\multicolumn{1}{c}{\textbf{SPL}$\uparrow$} & 
\multicolumn{1}{c}{TL} & \multicolumn{1}{c}{NE$\downarrow$} & \multicolumn{1}{c}{OSR$\uparrow$} & \multicolumn{1}{c}{\textbf{SR}$\uparrow$} & \multicolumn{1}{c}{\textbf{SPL}$\uparrow$} & 
\multicolumn{1}{c}{TL} & \multicolumn{1}{c}{NE$\downarrow$} & \multicolumn{1}{c}{OSR$\uparrow$} & \multicolumn{1}{c}{\textbf{SR}$\uparrow$} & \multicolumn{1}{c}{\textbf{SPL}$\uparrow$} \\
\midrule
\midrule
Human    
& --
& -- & -- & -- & -- & -- 
& -- & -- & -- & -- & --
& 11.85 & 1.61 & 90 & 86 & 76 \\
\midrule

Seq2Seq~\cite{r2r}
& --
& 11.33 & 6.01 & 53 & 39 & -- 
& 8.39 & 7.81 & 28 & 21 & -- 
& 8.13 & 7.85 & 27 & 20 & -- \\
RCM~\cite{wang2019reinforced}
& --
& 10.65 & 3.53 & 75 & 67 & -- 
& 11.46 & 6.09 & 50 & 43 & -- 
& 11.97 & 6.12 & 50 & 43 & 38 \\

Speaker Follower~\cite{sf}
& --
& -- & 3.36 & 74 & 66 & -- 
& -- & 6.62 & 45 & 36 & -- 
& 14.82 & 6.62 & - & 35 & 28  \\
EnvDrop~\cite{envdrop}
& --
& 11.00 & 3.99 & -- & 62 & 59
& 10.70 & 5.22 & -- & 52 & 48
& 11.66 & 5.23 & 59 & 51 & 47 \\

\midrule
\rowcolor{Cerulean!20}\multicolumn{17}{l}{\emph{VLN Specialists with Vision-Language-Action Pretraining:}}\\
PREVALENT~\cite{prevalent}
& --
& 10.32 & 3.67 & -- & 69 & 65
& 10.19 & 4.71 & -- & 58 & 53
& 10.51 & 5.30 & 61 & 54 & 51 \\
AirBert~\cite{guhur2021airbert}$\dag$
& --
& 11.09 & 2.68 & -- & 75 & 70
& 11.78 & 4.10 & -- & 62 & 56 
& 12.41 & 4.13 & -- & 62 & 57 \\
VLNBert~\cite{recurrent_vln}
& --
& 11.13 & 2.90 & -- & 72 & 68
& 12.01 & 3.93 & -- & 63 & 57
& 12.35 & 4.09 & 70 & 63 & 57 \\
MARVAL~\cite{newpath}$\dag$
& --
& 10.60 & 2.99 & -- & 73 & 69
& 10.15 & 4.06 & -- & 65 & 61
& 10.22 & 4.18 & 67 & 62 & 58 \\

HAMT~\cite{hamt}
& --
& 11.15 & 2.51 & -- & 76 & 72
& 11.46 & 2.29 & -- & 66 & 61
& 12.27 & 3.93 & 72 & 65 & 60 \\
HOP+~\cite{hop+}
& --
& 11.31 & 2.33 & -- & 78 & 73 
& 11.76 & 3.49 & -- & 67 & 61
& 12.67 & 3.71 & -- & 66 & 60 \\

DUET~\cite{duet}
& --
& 12.32 & 2.28 & 86 & 79 & 73 
& 13.94 & 3.31 & 81 & 72 & 60
& 14.73 & 3.65 & 76 & 69 & 59 \\
BEVBert~\cite{bevbert}
& --
& 13.56 & 2.17 & 88 & \textbf{81} & \textbf{74}
& 14.55 & 2.81 & 84 & \textbf{75} & \textbf{64}
& 15.87 & 3.13 & 81 & \textbf{73} & \textbf{62} \\

\midrule\midrule

\rowcolor{Cerulean!20}\multicolumn{17}{l}{\emph{Language Models zero-shot VLN:}}\\
NavGPT (GPT-4)~\cite{navgpt}
& \checkmark
& -- & -- & -- & -- & --
& 11.45 & 6.46 & 42 & 34 & 29
& -- & -- & -- & -- & -- \\
MapGPT (GPT-4)~\cite{mapgpt}
& \checkmark
& -- & -- & -- & -- & --
& -- & 6.92 & 58 & 39 & 26
& -- & -- & -- & -- & -- \\
DiscussNav (GPT-4)~\cite{discussnav}
& \checkmark
& -- & -- & -- & -- & --
& 9.69 & 5.32 & 61 & 43 & 40
& -- & -- & -- & -- & -- \\
\midrule
\rowcolor{Cerulean!20}\multicolumn{17}{l}{\emph{Langage Models with/as VLN Specialists:}}\\
NavCoT (LLaMA2-7B)~\cite{lin2024navcot}
& \xmark
& 10.08 & 6.46 & 48 & 41 & 38
& 9.95 & 6.26 & 48 & 40 & 37
& -- & -- & -- & -- & -- \\
LangNav (LLaMA2-7B)~\cite{pan2023langnav}$\dag$
& \xmark
& -- & -- & -- & 56 & --
& -- & -- & -- & 46 & --
& -- & -- & -- & -- & -- \\

NaviLLM (Vicuna-7B)~\cite{navillm}
& \xmark
& -- & -- & -- & -- & --
& 12.81 & 3.51 & -- & 67 & 59
& 13.21 & 3.71 & -- & 68 & 60 \\
NavGPT2 (FlanT5-XL-1.5B)~\cite{navgpt2}
& \checkmark
& 13.02 & 3.34 & 74 & 69 & 62
& 13.68 & 3.37 & 74 & 68 & 56
& -- & -- & -- & -- & -- \\
\quad w/ PREVALENT
& \checkmark
& 12.44 & 2.97 & 80 & 73 & 65
& 12.81 & 3.33 & 79 & 70 & 59
& 13.51 & 3.40 & 77 & 71 & 60 \\
NavGPT2  (FlanT5-XXL-5B)~\cite{navgpt2}
& \checkmark
& 13.08 & 2.98 & 79 & 74 & 65
& 13.25 & 3.18 & 80 & 71 & 60
& -- & -- & -- & -- & -- \\
\quad w/ PREVALENT
& \checkmark
& 14.13 & 2.84 & 83 & 74 & 63
& 14.01 & 2.98 & 84 & 74 & 61
& 14.74 & 3.33 & 80 & 72 & 60 \\
$\textbf{\ours}$  (\textbf{FlanT5-XL-1.5B}) 
& \checkmark
& 12.02 & 3.21 & 79 & 73 & 65
& 12.84 & 3.19 & 80 & 72 & 61
& -- & -- & -- & -- & -- \\
\quad w/ PREVALENT
& \checkmark
& 12.26 & 2.67 & 82 & \textbf{76} & \textbf{67}
& 13.33 & 2.81 & 83 & \textbf{75} & \textbf{63}
& 14.00 & 3.03 & 80 & \textbf{74} & \textbf{63} \\
\midrule
\bottomrule
\end{tabular}}
\label{tab:r2r_full}
\vspace{-15pt}
\end{table*}

\paragraph{Compared Baselines.} We have categorized the previous SOTA methods and listed them in a Tab.~\ref{tab:r2r_full}. Seq2Seq~\cite{r2r}, RCM~\cite{rcm}, Speaker Follower~\cite{sf}, and EnvDrop~\cite{envdrop} are early methods that typically use architectures like recurrent neural networks and were not trained beyond the navigation policy in MP3D. Later, some methods~\cite{prevalent,guhur2021airbert,recurrent_vln,newpath,hamt,hop+,duet,bevbert} employed a two-stage training process: first, pretraining an initialized vision-language model in the MP3D simulation, followed by fine-tuning the navigation policy with the vision-language model. These VLN specialists, utilizing their Vision-Language-Action pretraining method, achieve significant improvements over earlier approaches. Subsequently, methods using LLMs emerged. NavGPT~\cite{navgpt}, MapGPT~\cite{mapgpt} and DiscussNav~\cite{discussnav} attempt to complete the VLN task in a training-free manner, leveraging the zero-shot capability of GPT-4 and directly prompting the LLM to make decisions for the next action. Later, some methods~\cite{lin2024navcot,pan2023langnav,navillm} fine-tune LLMs as VLN experts, while NavGPT2~\cite{navgpt2} connects the LLM and VLN specialist through a latent space.  

\paragraph{Main Results.} Table~\ref{tab:r2r_full} presents the metrics of \textbf{\ours} compared to the methods mentioned above on the R2R benchmark. Compared to other LLM-leveraged methods, our approach achieves the best performance with the fewest parameters—1.5B. NaviLLM~\cite{navillm} is the first fully fine-tuned 7B LLM used as an action generator, and \ours outperforms it by $6\%$ SR and $3\%$ SPL on the test split. This indicates that maintaining the text generation capability of the LLM is crucial for navigation. Compared to NavGPT2, ATD outperforms its 1.5B model by $3\%$ in SR and $3\%$ in SPL on the test split. This suggests that our human-like left-right brain architecture is superior to the single-brain bridged architecture. BEVBert~\cite{bevbert} utilizes additional depth information to construct a bird-eye-view map. ATD achieves comparable performance on the validation unseen split and outperforms it on the test split. This indicates that our ATD using an LLM's imagination ability enhances the generalization of the navigation policy. 

\paragraph{Zero-shot Cross Environments evaluation.} We compared the cross-environment capability of ATD with that of traditional navigation policies, as shown in the Tab.~\ref{tab:zero_shot}. Both models were trained on R2R and evaluated on REVERIE~\cite{reverie}. It was observed that ATD outperformed by $4\%$ in OSR, $3\%$ in SR, and $4\%$ in SPL. This demonstrates that our model has better generalization ability. 
\begin{figure*}[t]
    \centering
    \begin{minipage}[b]{0.67\linewidth}
        \centering
        \includegraphics[width=\linewidth]{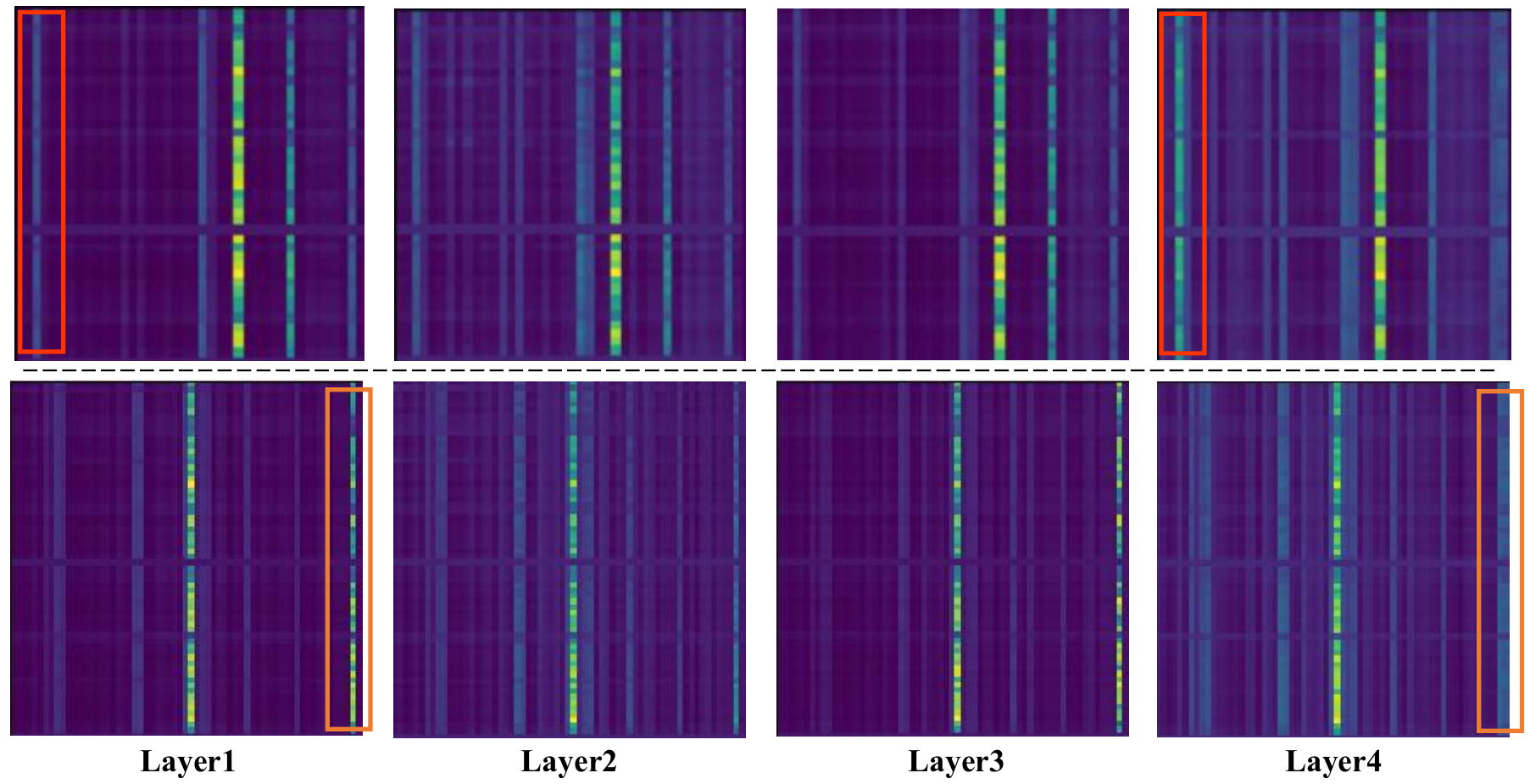}
        \caption{\textbf{Visualization of attention matrices across four SGCA layers.} \textbf{Row 1}: The red box shows the increasing emphasis on important information, while \textbf{Row 2}: The orange box illustrates the suppression of completed navigation steps.}
        \label{fig:attention}
    \end{minipage}
    \hfill
    \begin{minipage}[b]{0.32\linewidth}
        \centering
        \includegraphics[width=\linewidth]{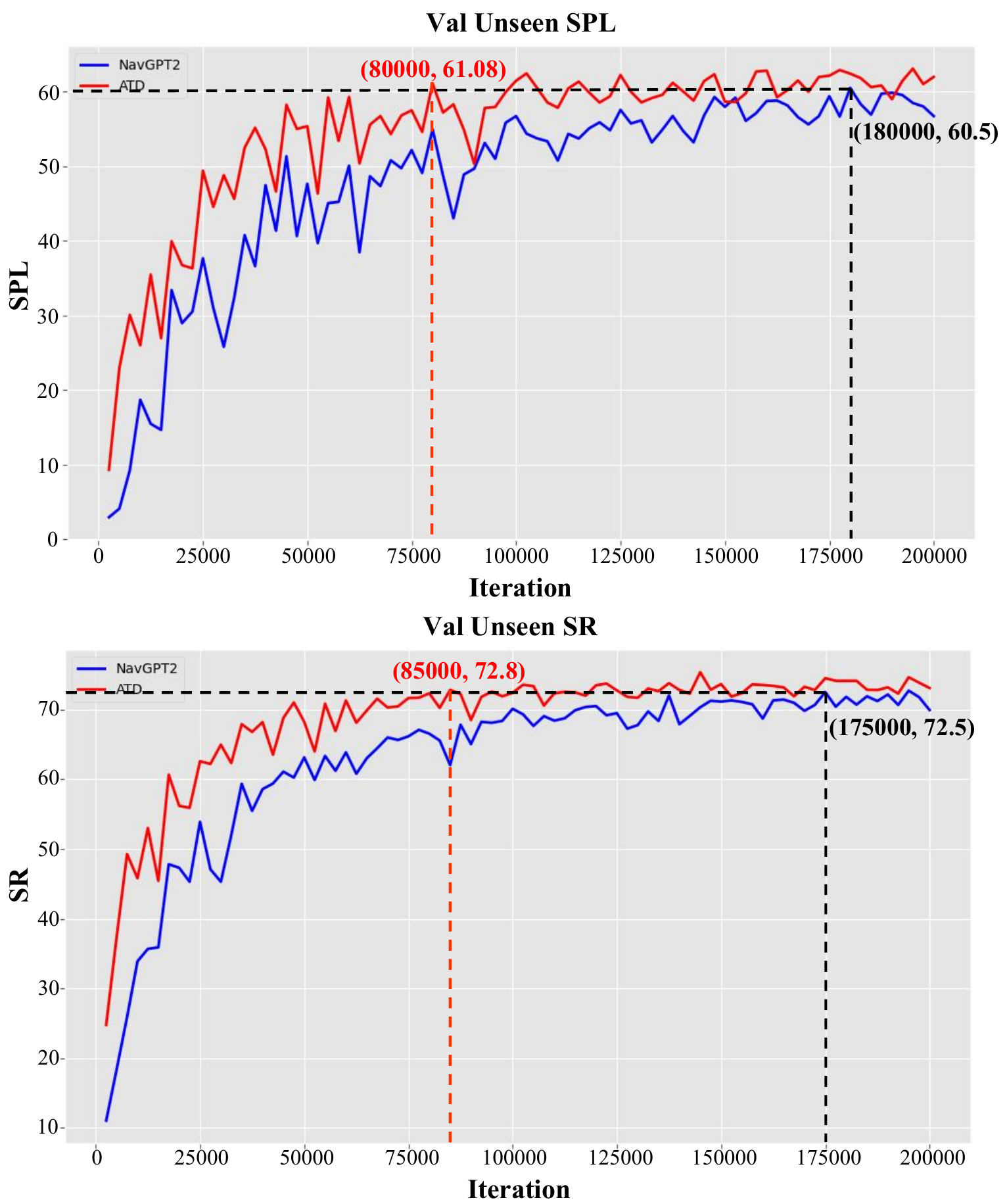}
        \vspace{-20pt}
        \caption{\textbf{Comparison of convergence speeds.} The red vertical lines highlight the points where ATD surpasses NavGPT2's maximum SPL and SR values.}
        \label{fig:speed}
    \end{minipage}
\end{figure*}

\subsection{Qualitative Results} \label{sec:exp_qualitative}

\paragraph{Attention Visualization.} As shown in Fig.~\ref{fig:attention}, we visualize the attention matrices $\mathbf{A}$ produced by the SGCA at each layer. For this visualization, the number of SGCA layers is set to 4. As shown in \textbf{Row 1} of Fig.~\ref{fig:attention}, a region within the red box gradually gains emphasis during the information interaction between state estimation and imagination. By the final layer (Layer 4), its attention weight becomes significantly higher compared to earlier layers. This suggests that SGCA may indeed play a role in filtering and highlighting the more important information within the imagination. In \textbf{Row 2}, the region within the orange box shows a noticeable decrease in attention weight as it propagates through the layers, becoming minimal in the final output. This may indicate that, under the supervision of state estimation, SGCA learns to suppress certain parts of the information from imagination—potentially those related to navigation steps that have already been completed. 

\paragraph{Comparison of Convergence Speed.} Line charts of the SR and SPL metrics on the validation unseen split over training iterations are shown in Fig.~\ref{fig:speed}. This figure allows us to compare the convergence speeds of NavGPT2 and \ours. The chart above in Fig.~\ref{fig:speed} shows the variation in the SPL curve. The black line highlights the point $(180000, 60.5)$, which represents the maximum SPL value for NavGPT2, indicating its corresponding coordinates. On the left side, observing the red vertical line, we can see that ATD surpasses the maximum SPL value of NavGPT2 at around $80,000$ iterations, reaching this point $100,000$ iterations faster. Similarly, in the line chart below of Fig.~\ref{fig:speed}, by observing the position of the red vertical line corresponding to the black line, we can see that ATD reaches NavGPT2's best SR metric $90,000$ iterations faster.

\begin{table*}[t]
\centering
\begin{minipage}[t]{0.25\textwidth}
\centering
\caption{Compare zero-shot result on REVERIE.}
\resizebox{\textwidth}{!}{
\definecolor{Gray}{gray}{0.94}
\begin{tabular}{lccccc}
\toprule
\multicolumn{1}{c}{\multirow{2}{*}{\textbf{Methods}}} & 
\multicolumn{3}{c}{\textbf{REVERIE}} \\

\cmidrule(r){2-4}
\multicolumn{1}{c}{} &
\multicolumn{1}{c}{OSR$\uparrow$} &
\multicolumn{1}{c}{SR$\uparrow$} &
\multicolumn{1}{c}{SPL$\uparrow$} \\

\midrule
DUET
& 29 & 24 & 19   \\
$\text{\ours}_{\text{FlanT5-XXL}}$
& \textbf{33} & \textbf{27} & \textbf{23} \\
\midrule
\end{tabular}
}
\label{tab:zero_shot}
\end{minipage}
\hfill
\begin{minipage}[t]{0.74\textwidth}
\centering
\caption{\textbf{Comparison of different LLMs.}}
\resizebox{\textwidth}{!}{
\definecolor{Gray}{gray}{0.94}
\begin{tabular}{llccccccccc}
\toprule
\multicolumn{1}{c}{\multirow{2}{*}{Methods}} & 
\multicolumn{5}{c}{\textbf{\texttt{Val Seen}}} &
\multicolumn{5}{c}{\textbf{\texttt{Val Unseen}}} \\

\cmidrule(r){2-6}
\cmidrule(r){7-11}

\multicolumn{1}{c}{} &
\multicolumn{1}{c}{TL} &
\multicolumn{1}{c}{NE$\downarrow$} &
\multicolumn{1}{c}{OSR$\uparrow$} &
\multicolumn{1}{c}{\textbf{SR}$\uparrow$} &
\multicolumn{1}{c}{\textbf{SPL}$\uparrow$} &
\multicolumn{1}{c}{TL} & \multicolumn{1}{c}{NE$\downarrow$} & \multicolumn{1}{c}{OSR$\uparrow$} & \multicolumn{1}{c}{\textbf{SR}$\uparrow$} & \multicolumn{1}{c}{\textbf{SPL}$\uparrow$} \\

\midrule
$\text{\ours}_{\text{FlanT5-XL}}$
& 12.26 & 2.67 & 81.78 & \textbf{75.61} & 67.49
& 13.33 & 2.81 & 82.76 & \textbf{74.63} & 63.08 \\
$\text{\ours}_{\text{FlanT5-XXL}}$
& 11.38 & 2.47 & 80.51 & 75.51 & \textbf{70.02}
& 11.91 & 2.82 & 81.01 & 73.82 & \textbf{65.35} \\
\midrule
\end{tabular}
}
\label{tab:LLM_compare}
\end{minipage}
\end{table*}

\begin{table}[t]
\centering
\begin{minipage}{0.52\textwidth}
    \caption{\textbf{Ablation of SELLM and IMLLM.} We investigate the improvement of SELLM and IMLLM on the model performance based on the baseline.}
    \vspace{0.2em}
    \small
    \resizebox{\textwidth}{!}{
    \begin{tabular}{ccc|ccccc}
    \toprule
     & \textbf{SEM} & \textbf{IM} &TL &NE $\downarrow$ & OSR $\uparrow$ & \textbf{SR $\uparrow$} & \textbf{SPL $\uparrow$} \\ 
    \midrule
    \multirow{4}{*}{\rotatebox{90}{\textbf{\texttt{\parbox{1cm}{\centering Val\\Seen}}}}} &  &  &  12.14 & 3.45 & 76 & 68 & 62\\
    & \checkmark &  &  13.55 & 3.12 & 79 & 71 & 63\\
     & & \checkmark & 13.40 & 3.15 & 77 & 71 & 62\\
     & \checkmark & \checkmark & 12.26 & \textbf{2.67} & \textbf{82} & \textbf{76} & \textbf{67}\\
     
     \midrule
     
    \multirow{4}{*}{\rotatebox{90}{\textbf{\texttt{\parbox{1cm}{\centering Val\\Unseen}}}}} & &  & 13.70 & 4.20 & 72 & 63 & 52\\
    & \checkmark &  & 13.23 & 3.05 & 80 & 72 & 61 \\
     & & \checkmark & 13.87 & 2.97 & 80 & 73 & 60 \\
     & \checkmark & \checkmark & 13.33 & \textbf{2.81} & \textbf{83} & \textbf{75} & \textbf{63}\\
     \bottomrule
    \end{tabular}
    }
    \label{tab:ablation}
\end{minipage}
\hfill
\begin{minipage}{0.47\textwidth}
    \centering
    \caption{\textbf{Ablation of SGCA layer numbers}.}
    \small
    \resizebox{\textwidth}{!}{
    \begin{tabular}{c r c c c c }
    \toprule
    \multirow{2}{*}{} & \multirow{2}{*}{\textbf{Metrics}}  & \multicolumn{4}{c}{\textbf{Number of Layers}} \\ \cmidrule(lr){3-6} &  & 1 & 2 & 3 & 4 \\

    \midrule
    \multirow{4}{*}{\rotatebox{90}{\textbf{\texttt{\parbox{1cm}{\centering Val\\Seen}}}}} 
    &  NE $\downarrow$  & 3.05 & 2.63 & 2.88 & 2.67  \\
    &  OSR $\uparrow$  & 78.35 & 82.47 & 80.8 & 81.78  \\ 
    &  \textbf{SR} $\uparrow$  & 71.69 & 75.91 & \textbf{76.3} & 75.61  \\ 
    &  \textbf{SPL} $\uparrow$  & 63.15 & 67.46 & 67.13 & \textbf{67.49}  \\ 

    \midrule

    \multirow{4}{*}{\rotatebox{90}{\textbf{\texttt{\parbox{1cm}{\centering Val\\Unseen}}}}} 
    &  NE $\downarrow$  & 2.82 & 2.9 & 2.88 & 2.81  \\
    &  OSR $\uparrow$  & 82.08 & 82.12 & 82.12 & 82.76  \\ 
    &  \textbf{SR} $\uparrow$  & 74.54 & 73.95 & \textbf{74.93} & 74.63  \\ 
    &  \textbf{SPL} $\uparrow$  & 63.15 & 63.05 & \textbf{63.91} & 63.08  \\

    \bottomrule
    \end{tabular}
    }
    \label{tab:num_layers}
\end{minipage}
\end{table}

\subsection{Ablation Study} \label{sec:exp_abla}

\paragraph{Effect of the LLM model.} To demonstrate the transferability of our method, we also implemented the ATD approach on a different LLM, as shown in the Tab.~\ref{tab:LLM_compare}. FlanT5-XL has 1.5 billion parameters, while FlanT5-XXL has 5 billion. It can be observed that $\text{\ours}_{\text{FlanT5-XL}}$ achieves slightly higher SR on both \texttt{val seen} and \texttt{val unseen} sets, whereas $\text{\ours}_{\text{FlanT5-XLL}}$ shows a significant advantage in SPL. This might be because $\text{\ours}_{\text{FlanT5-XXL}}$ has higher confidence in reaching the destination and thus stops early, resulting in a higher SPL. However, if the decision is incorrect, it could lead to a lower SR.

\paragraph{Effectiveness of State Estimation LLM and Imagination LLM.} As shown in Tab.~\ref{tab:ablation}, we investigate the effectiveness of two key modules in our method. When both State Estimation LLM (SELLM) and Imagination LLM (IMLLM) are removed, it corresponds to our baseline model, which is DUET without the local branch and without BERT pretraining in the MP3D simulation. Specifically, including either the SELLM or IMLLM module leads to a significant improvement over the baseline. However, the best performance is achieved when both modules interact, as seen in the full human-like left-right brain ATD. This demonstrates the effectiveness of both our approach and its structural design. Our results show that the full model outperforms others on both the \textbf{\texttt{val seen}} and \textbf{\texttt{val unseen}} datasets. Compared to the baseline, ATD shows an $8\%$ increase in SR and a $5\%$ increase in SPL, demonstrating strong performance in key navigation metrics. These improvements highlight the power of combining SELLM and IMLLM, underscoring the importance of both modules working together for optimal task performance.

\paragraph{Effect of different SGCA Layer numbers for features fusion.} Tab.~\ref{tab:num_layers} shows the performance of ATD with SGCA layer from 1 to 4. For both \textbf{SR} and \textbf{SPL} metrics, the best performance is achieved when the number of layers is either 3 or 4. In \textbf{\texttt{val seen}}, increasing the number of SGCA layers significantly improves performance: SR from $71.69\%$ to $76.3\%$ and SPL from $63.14\%$ to $67.49\%$. However, in \textbf{\texttt{val unseen}}, the improvement from increasing the number of layers is marginal. Surprisingly, when the number of layers is set to 1, the model performs much better on \textbf{\texttt{val unseen}} than on \textbf{\texttt{val seen}}. This suggests that ATD has a relatively high lower bound for generalization.

\section{Conclusion}
\label{sec:conclusion}

In this work, we have presented the Adaptive Text Dreamer (ATD), a novel dual-branch vision-language navigation framework that leverages large language models to perform dynamic and adaptive imagination through language-based reasoning. By integrating a left-brain State Estimation LLM and a right-brain Imagination LLM, our approach effectively addresses the challenge of partial observability in Vision-and-Language Navigation, enabling agents to anticipate critical and complex semantic cues beyond their immediate perception. The latent adaptive cross-interactive imagination embedding bridges the language-based imagination with a graph-based navigation policy, resulting in improved decision-making. Experimental results on the R2R benchmark demonstrate that ATD not only outperforms previous state-of-the-art methods but also achieves these gains with fewer parameters and lower computational costs. Our study highlights the potential of leveraging advanced linguistic abstractions for targeted imagination in embodied AI, paving the way for more efficient and semantically grounded navigation agents.

\maketitlesupplementary
\begin{abstract}
    \noindent This supplementary material expands upon our main study by providing additional details and data to enhance the reproducibility of our proposed method \ours. \\
    \noindent $\triangleright$ \textbf{Sec.~\ref{sec:data collect}}: Configuration of Data Collection Progress for State Estimation LLM and Imagination LLM. This includes the design of prompts, collection of candidate node captions, and data visualization. \\
    \noindent $\triangleright$ \textbf{Sec.~\ref{sec:more implementation}}: This section offers a detailed elucidation of the metrics, delves into the intricacies of the training process, and presents a comprehensive introduction to the comparison methods. \\
    \noindent $\triangleright$ \textbf{Sec.~\ref{sec:more experiment}}: Additional experiments are designed and presented to investigate the effectiveness of the model. \\
    \noindent $\triangleright$ \textbf{Sec.~\ref{sec:discussion}}: Discusses the limitations of our work and explores prospects for future research, also discusses the broader effect of our work.
\end{abstract}
\section{Instruction Data Collection}
\label{sec:data collect}
To train the State Estimation LLM and Imagination LLM, we separately collect instruction data for Q-Former fine-tuning of each LLM. 

\begin{figure*}[h]
\centering
\includegraphics[width=0.99\linewidth]{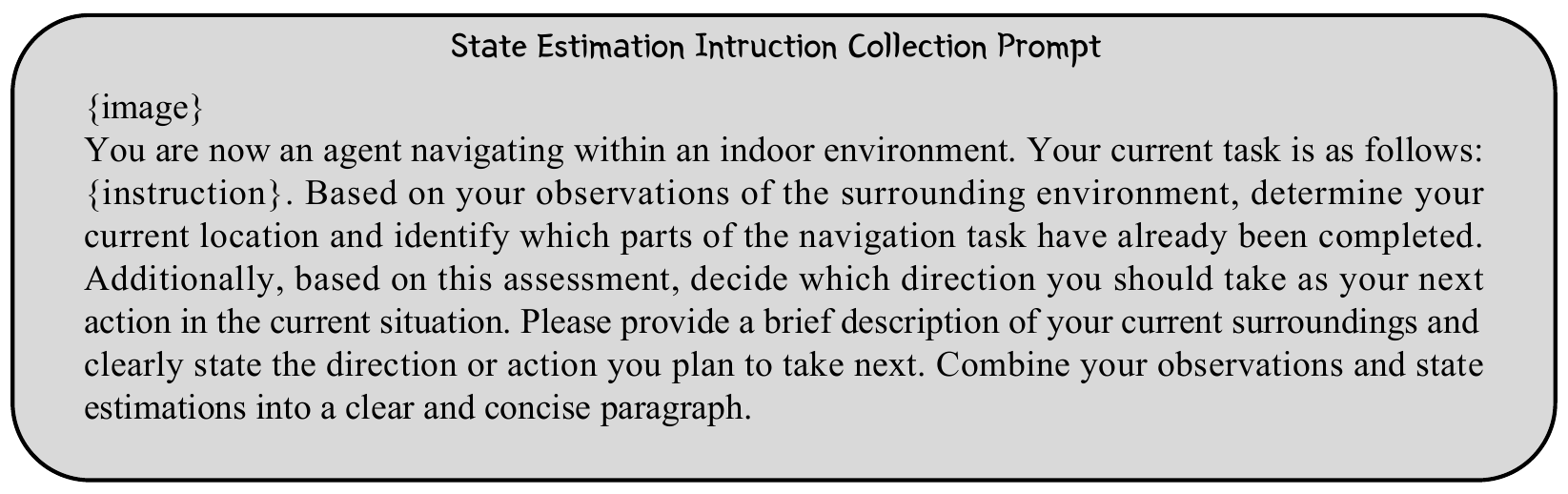}
\vspace{-10pt}
\caption{\textbf{State Estimation generation prompt.}}
\label{fig:state prompt}
\end{figure*}

\noindent \textbf{State Estimation Intruction Collection.} We employ a more capable LLM~\cite{achiam2023gpt} to perform text-based state estimation of the navigation process and use its outputs as ground truth to train our left brain. This process is more akin to distilling the state estimation capability into the Q-Former that we fine-tune. The prompt is shown in Fig.~\ref{fig:state prompt}.

\begin{figure*}[t]
\centering
\includegraphics[width=0.99\linewidth]{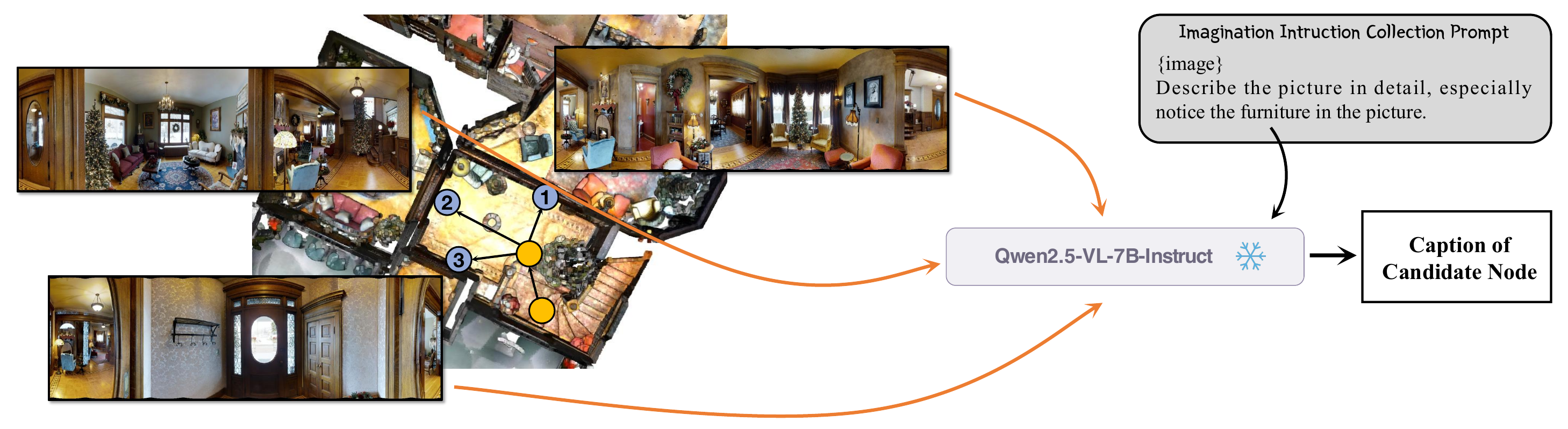}
\vspace{-10pt}
\caption{\textbf{Imagination Ground Truth Collection.}}
\label{fig:imagination collection}
\end{figure*}

\noindent \textbf{Imagination Intruction Collection.} To train the Imagination LLM, we gather the captions of candidate nodes associated with each node to serve as the training ground truth. An illustration of this collection process is shown in Fig.~\ref{fig:imagination collection}. As depicted, for each current node, we first collect images from all candidate nodes and stitch them together into a panoramic view. We then employ the Qwen2.5-VL-7B-Instruct~\cite{qwen25} model to generate a caption for the panoramic image. By combining these captions, we produce the imagination ground truth for the current node. 

\begin{figure*}[t]
\centering
\includegraphics[width=0.99\linewidth]{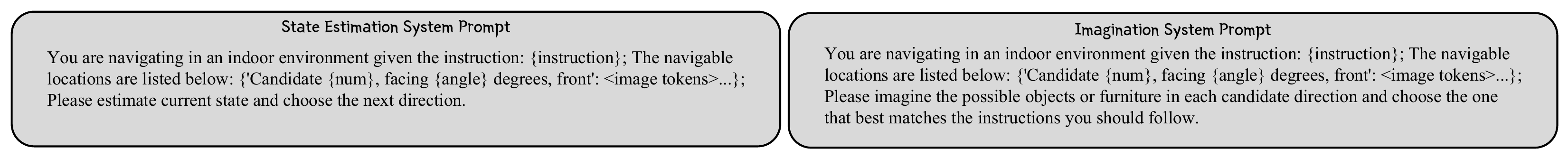}
\vspace{-10pt}
\caption{\textbf{State Estimation System Prompt and Imagination System Prompt.}}
\label{fig:system prompt}
\end{figure*}

\noindent \textbf{System Prompt.} As described in our method section, both the Left Brain and Right Brain use prompts to integrate the visual information generated by the Q-Former with the instruction input before passing it to the LLM. The system prompts for the State Estimation LLM and the Imagination LLM are illustrated in the Fig.~\ref{fig:system prompt}. 

\section{Additional Implementation Details}
\label{sec:more implementation}

\subsection{Evaluation Metrics}
In this section, we present a comprehensive overview of the evaluation metrics employed in our study. Consistent with prior research, we utilize five key metrics: Trajectory Length (\textbf{TL}), Navigation Error (\textbf{NE}), Success Rate (\textbf{SR}), Oracle Success Rate (\textbf{OSR}), and Success Weighted by Path Length (\textbf{SPL}). The following subsections provide a detailed description of each metric along with its corresponding mathematical formulation: 

\begin{itemize}[leftmargin=*]
    \item \textbf{TL.} The total length of the predicted trajectory, calculated as the sum of the shortest path distances between consecutive viewpoints along the trajectory.
    \begin{equation}
        L_{\text{traj}} = \sum_{i=1}^{N-1} d(v_i, v_{i+1}),
    \end{equation}
    where $v_i$ is the $i$-th viewpoint in the predicted path, $d(a,b)$ is the shortest path distance between viewpoints $a$ and $b$, and $N$ is the number of viewpoints.
    
    \item \textbf{NE.} The shortest path distance between the final viewpoint of the predicted trajectory and the goal viewpoint.
    \begin{equation}
    E_{\text{nav}} = d(v_N, v^*),
    \end{equation}
    where $v_N$ is the last viewpoint in the predicted trajectory and $v^*$ is the goal viewpoint.
    
    \item \textbf{SR.} Indicates whether the agent successfully reached the goal. For tasks with a set of goal viewpoints, success is achieved if the final viewpoint of the predicted path is within the goal set. Otherwise, success is defined by whether the navigation error is below a certain threshold $\epsilon$.
    \begin{equation}
        S = 
            \begin{cases}
            1 & \text{if } v_N \in G \\
            0 & \text{otherwise,}
            \end{cases}
    \end{equation}
    or, if no goal set exists:
    \begin{equation}
        S = 
            \begin{cases}
            1 & \text{if } d(v_N, v^*) < \epsilon \\
            0 & \text{otherwise,}
            \end{cases}
    \end{equation}      
    where $v_N$ is the final viewpoint of the predicted path, $G$ is the goal viewpoint set, $v^*$ is the goal viewpoint, and $\epsilon$ is the success threshold.
    
    \item \textbf{OSR.} Measures whether any viewpoint along the predicted path falls within the goal set, reflecting the best possible success if the agent stopped at the closest point to the goal.
    \begin{equation}
        S_{\text{oracle}} = 
            \begin{cases}
            1 & \text{if } \exists v_i \in G \\
            0 & \text{otherwise,}
            \end{cases}
    \end{equation}  
    or, if no goal set:
    \begin{equation}
        S_{\text{oracle}} = 
            \begin{cases}
            1 & \text{if } \min_i d(v_i, v^*) < \epsilon \\
            0 & \text{otherwise,}
            \end{cases}
    \end{equation}  
    where $v_i$ is the $i$-th viewpoint along the path.

    \item \textbf{SPL.} A metric combining success and efficiency, penalizing longer trajectories relative to the shortest possible path length.
    \begin{equation}
        \text{SPL} = S \times \frac{L_{\text{gt}}}{\max(L_{\text{traj}}, L_{\text{gt}})},
    \end{equation}
    where $L_{\text{gt}}$ is the ground-truth path length, $L_{\text{traj}}$ is the predicted path length.  

\end{itemize}

\subsection{Training Details}

Following NavGPT2~\cite{navgpt2}, our best-performing model is trained with additional synthetic data from PREVALENT~\cite{prevalent}. We conducted an ablation study by excluding the PREVALENT data and observed that this synthetic data is vital in preventing our method from overfitting. Without incorporating the synthetic data, the validation loss plateaus prematurely during the early stages of training. The parameter size of our model is 1.5B because only the encoder was used during policy training, resulting in half of the parameters of the Flant5-XL model.

\subsection{Compared Baselines}
In this section, we provide a detailed overview of comparative methods, elaborating on their distinct methodological designs.  
\begin{itemize}[leftmargin=*]
    \item \textbf{Seq2Seq~\cite{r2r}.} The study introduces the Room-to-Room (R2R) dataset, the first benchmark for visually-grounded natural language navigation. It consists of real building layouts and natural-language instructions that guide agents between rooms, providing a realistic setting to evaluate models on understanding spatial relationships and visual contexts. And this paper presenting a framework for robots to interpret natural-language navigation instructions using visual inputs. The core method leverages sequence-to-sequence translation techniques, treating the task as a visually grounded translation problem akin to Visual Question Answering. To enable research, the authors develop the Matterport3D Simulator, a large-scale reinforcement learning environment built on real-world imagery, allowing robots to learn navigation policies in simulated 3D environments.

    \item \textbf{RCM~\cite{rcm}.} Reinforced Cross-Modal Matching employs reinforcement learning to enforce both local (via a reasoning navigator for visual-scene grounding) and global (via a matching critic providing intrinsic rewards for instruction-trajectory alignment) cross-modal alignment. 

    \item \textbf{Speaker Follower~\cite{sf}.} The approach addresses data scarcity and reasoning challenges via: (1) it synthesizes new natural language instructions for data augmentation and (2) implements pragmatic reasoning to evaluate how well candidate action sequences explain given instructions. 

    \item \textbf{EnvDrop~\cite{envdrop}.} This paper presents a two-stage training framework. In the first stage, the agent is trained via a combination of imitation learning and reinforcement learning, integrating the advantages of off-policy and on-policy optimization. The second stage involves fine-tuning with "unseen" triplets (environment, path, instruction), generated using an "environmental dropout" method that mimics unseen environments by randomly removing objects to enhance variability. 

    \item \textbf{PREVALENT~\cite{prevalent}.} This paper presents a pre-training and fine-tuning paradigm for VLN tasks. The proposed agent, Prevalent, is pre-trained via self-supervised learning on a large dataset of image-text-action triplets to learn generic representations of visual environments and language instructions. It can be integrated into existing VLN frameworks as a plug-and-play module. And we use it in our main experiments. 

    \item \textbf{AirBert~\cite{guhur2021airbert}.} The core methods include: 1) constructing BnB, a large-scale dataset by collecting image-caption pairs (IC pairs) from hundreds of thousands of online rental listings and generating millions of path-instruction pairs (PI pairs) via automatic strategies; 2) proposing a shuffling loss to enhance learning of temporal order in PI pairs. 

    \item \textbf{VLNBert~\cite{recurrent_vln}.} This paper proposes a recurrent Vision-Language BERT model, equipping BERT with a recurrent function to maintain cross-modal state information and address the history-dependent decision-making challenge in the partially observable Markov decision process. 

    \item \textbf{MARVAL~\cite{newpath}.} This paper presents a novel approach to scaling VLN by leveraging synthetic instructions and imitation learning. The authors construct navigation trajectories across 500+ indoor environments with densely-sampled 360° panoramas, then generate 4.2 million visually-grounded instruction-trajectory pairs using Marky, a multilingual instruction generator, and synthesize novel viewpoint images via image-to-image GAN. This work highlights the power of synthetic data augmentation and imitation learning for advancing instruction-following agents in VLN.

    \item \textbf{HAMT~\cite{hamt}.} HAMT employs a hierarchical ViT to encode past panoramic observations by modeling image-level features, spatial relations within panoramas, and temporal dynamics across historical panoramas. It fuses text instructions, historical context, and current observations to predict actions. The model is first pre-trained on proxy tasks (e.g., single-step action prediction, spatial relation prediction) and then optimized via reinforcement learning. 

    \item \textbf{HOP+~\cite{hop+}.} The method introduces three VLN-specific pre-training tasks: Action Prediction with History (APH) to leverage historical visual trajectories for action prediction, Trajectory Order Modeling (TOM) and Group Order Modeling (GOM) to enhance temporal order reasoning. Additionally, a memory network is designed for fine-tuning to dynamically select and summarize historical information, addressing representation inconsistency between pre-training and fine-tuning stages. The experiment validating the effectiveness of history-enhanced and order-aware learning. 

    \item \textbf{DUET~\cite{duet}.} The method dynamically combines fine-scale encoding of local visual observations with coarse-scale encoding on a globally built topological map via graph transformers, enabling joint long-term action planning and fine-grained cross-modal understanding. 

    \item \textbf{BEVBert~\cite{bevbert}.} The method constructs a hybrid map system, including a local metric map to explicitly aggregate incomplete observations and remove duplicates, and a global topological map to model navigation dependencies, balancing short-term reasoning and long-term planning. It pre-trains a multimodal map representation to enhance spatial-aware cross-modal reasoning. 

    \item \textbf{NavGPT~\cite{navgpt}.} The method enables NavGPT to perform zero-shot sequential action prediction by taking textual descriptions of visual observations, navigation history, and future explorable directions as inputs to reason about the agent's status and decide actions. It demonstrates abilities in decomposing instructions into sub-goals, integrating commonsense knowledge, identifying landmarks, tracking progress, and adjusting plans. The work leverages LLMs without specific training on navigation datasets, showcasing their potential for explicit reasoning in embodied scenes, though performance on zero-shot R2R tasks lags behind trained models. 

    \item \textbf{MapGPT~\cite{mapgpt}.} It introduces an online linguistic-formed map incorporating node information and topological relationships into prompts to help GPT understand the spatial environment, and proposes an adaptive planning mechanism for multi-step path planning based on the map. 

    \item \textbf{DiscussNav~\cite{discussnav}.} It employs large models with distinct capabilities as domain experts (e.g., for instruction understanding, environment perception, and completion estimation). The navigation agent actively discusses with these experts at each step to gather critical information before moving, correcting errors and filtering inconsistent decisions.

    \item \textbf{NavCoT~\cite{lin2024navcot}.} This paper employs Chain-of-Thought to enhance LLM-based VLN. At each timestep, the LLM acts as a world model to imagine the next observation based on instructions, selects the most aligned candidate observation, and determines actions through disentangled reasoning. Formalized training labels are constructed to guide the LLM in generating reasonable chain-of-thought outputs for improved action decisions. 

    \item \textbf{LangNav~\cite{pan2023langnav}.} It employs off-the-shelf vision systems to convert egocentric panoramic views into text descriptions and fine-tunes a pretrained language model to select actions based on these descriptions and trajectory history. 

    \item \textbf{NaviLLM~\cite{navillm}.} This paper presents NaviLLM, the first generalist model for embodied navigation, which adapts large language models (LLMs) to this field via schema-based instructions. This approach flexibly transforms various tasks into generation problems to unify diverse tasks and integrates multi-dataset sources (e.g., CVDN, SOON, ScanQA) during training. 

    \item \textbf{NavGPT2~\cite{navgpt2}.} The method integrates a frozen LLM with a topological graph-based navigation policy network. Visual observations are encoded into image tokens via a Q-former and fused with instruction text tokens to generate navigational reasoning. The navigation policy uses graph memory to model spatial structures, enabling long-term history tracking and backtracking. Trained in two stages, NavGPT-2 first undergoes visual instruction tuning on 10K navigation reasoning data generated by GPT-4V, then fine-tunes the policy network with R2R and PREVALENT datasets.

\end{itemize}

\section{Additional Experiment Results}
\label{sec:more experiment}

\noindent \textbf{More Cross Environment Experiment.} To validate the generalization capability of our model, we conduct zero-shot evaluation on additional datasets REVERIE and R4R, and compare our model with the DUET method, please see Tab.~\ref{tab:supp_zero}. Our ATD model outperforms on both the R4R and REVERIE datasets, demonstrating its strong generalizability.

\begin{table*}[h]
\centering
\caption{Comparison of zero-shot performance on R4R and REVERIE \textbf{\texttt{val unseen}} split.}
\resizebox{0.9\textwidth}{!}{
\definecolor{Gray}{gray}{0.94}
\begin{tabular}{llcccccccccc}
\toprule
\multicolumn{1}{c}{\multirow{2}{*}{Methods}} & 
\multicolumn{5}{c}{R4R} &
\multicolumn{3}{c}{REVERIE} \\

\cmidrule(r){2-6}
\cmidrule(r){7-9}

\multicolumn{1}{c}{} &
\multicolumn{1}{c}{nDTW$\uparrow$} &
\multicolumn{1}{c}{sDTW$\uparrow$} &
\multicolumn{1}{c}{OSR$\uparrow$} &
\multicolumn{1}{c}{\textbf{SR}$\uparrow$} &
\multicolumn{1}{c}{\textbf{SPL}$\uparrow$} &
\multicolumn{1}{c}{OSR$\uparrow$} &
\multicolumn{1}{c}{\textbf{SR}$\uparrow$} &
\multicolumn{1}{c}{\textbf{SPL}$\uparrow$} \\

\midrule
DUET
& 40.98 & 11.16 & 53.69 & 17.73 & 15.76  
& 28.94 & 24.42 & 19.09  \\
$\text{\ours}_{\text{FlanT5-XXL}}$
& \textbf{44.35} & \textbf{13.58} & \textbf{55.36} & \textbf{19.95} & \textbf{18.31} 
& \textbf{32.52} & \textbf{27.04} & \textbf{22.51} \\
\midrule
\end{tabular}
}

\label{tab:supp_zero}
\end{table*}

\noindent \textbf{Ablation of Left-right Brain Structure.} In Tab.~\ref{tab:supp_ablation}, we perform an ablation study on the left-right brain structure of \textbf{\emph{Cross from Left to Right}}. In \emph{Raw 2}, \textbf{\emph{Cross from Right to Left}} designates State Estimation as both the key and value and Imagination as the query, altering their roles in \textbf{SGCA} layers compared to ATD. In \emph{Raw 3}, the \textbf{\emph{Parallel Structure}} refers to the exclusion of the \textbf{SGCA} layer, where State Estimation and Imagination are summed in parallel and then fed into the subsequent policy for training. As can be seen from the results in the Tab.~\ref{tab:supp_ablation}, the \textbf{SGCA} structure in ATD that uses State Estimation to ground Imagination achieves the best performance (\emph{Raw 1}), which demonstrates the effectiveness of our \textbf{SGCA} layer design.

\begin{table*}[h]
\centering
\caption{Ablation of the left-right brain structure to validate the effectiveness of the SGCA layer.}
\resizebox{1.0\textwidth}{!}{
\definecolor{Gray}{gray}{0.94}
\begin{tabular}{llcccccccccc}
\toprule
\multicolumn{1}{c}{\multirow{2}{*}{Methods}} & 
\multicolumn{5}{c}{\textbf{\texttt{Val Seen}}} &
\multicolumn{5}{c}{\textbf{\texttt{Val Unseen}}} \\

\cmidrule(r){2-6}
\cmidrule(r){7-11}

\multicolumn{1}{c}{} &
\multicolumn{1}{c}{TL} &
\multicolumn{1}{c}{NE$\downarrow$} &
\multicolumn{1}{c}{OSR$\uparrow$} &
\multicolumn{1}{c}{SR$\uparrow$} &
\multicolumn{1}{c}{SPL$\uparrow$} &
\multicolumn{1}{c}{TL} & \multicolumn{1}{c}{NE$\downarrow$} & \multicolumn{1}{c}{OSR$\uparrow$} & \multicolumn{1}{c}{SR$\uparrow$} & \multicolumn{1}{c}{SPL$\uparrow$} \\

\midrule
ATD
& 12.26 & 2.67 & 81.78 & \textbf{75.61} & \textbf{67.49}
& 13.33 & 2.81 & 82.76 & \textbf{74.63} & \textbf{65.35} \\
Change to \textbf{\emph{Cross from Right to Left}}
& 12.38 & 3.17 & 76.79 & 71.79 & 63.97
& 12.14 & 2.92 & 79.65 & 73.95 & 63.58 \\
Change to \textbf{\emph{Parallel Structure}}
& 13.86 & 2.84 & 81.88 & 73.75 & 65.25
& 13.94 & 2.83 & 82.46 & 74.41 & 61.89 \\
\midrule
\end{tabular}
}

\label{tab:supp_ablation}
\end{table*}

\begin{figure*}[h]
\centering
\includegraphics[width=0.65\linewidth]{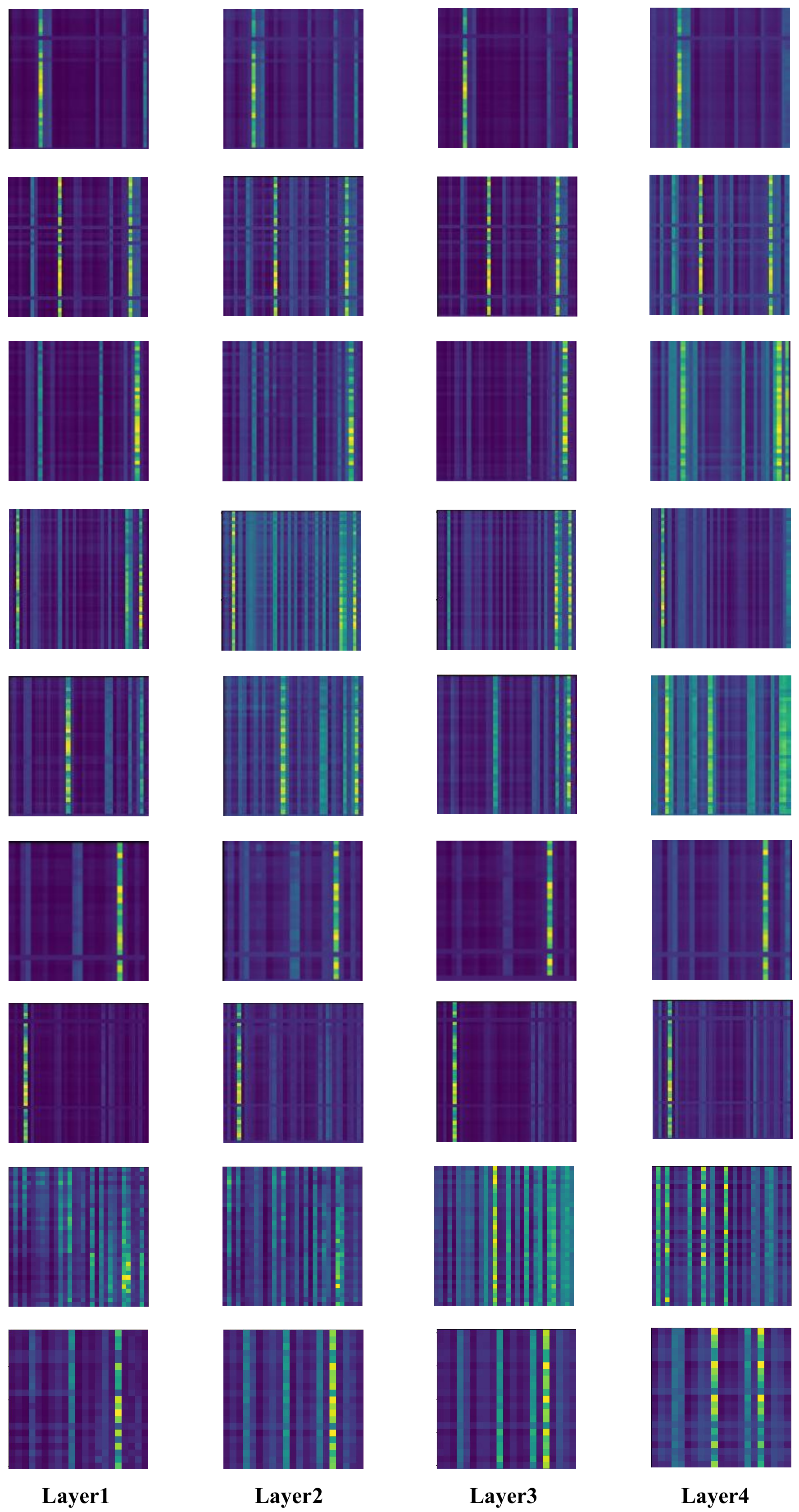}
\vspace{-5pt}
\caption{\textbf{Visualization of attention matrix for every layer.}}
\label{fig:supp attention}
\end{figure*}

\noindent \textbf{More Visualization of Attention Matrix.} As illustrated in Fig.~\ref{fig:supp attention}, we visualized additional attention matrices.

\section{Discussion}
\label{sec:discussion}
\noindent \textbf{Limitation and Future Work.} Currently, the data collected for training the imagination LLM is limited to candidate nodes one step ahead of the current node. This may restrict the model’s ability to perform long-horizon imagination. Future work could explore incorporating long-horizon imagination capabilities to more fully leverage the potential of the LLM’s imaginative reasoning.

\noindent \textbf{Broader Impact.} This work contributes to the field of vision-language navigation (VLN) by proposing an adaptive, language-based imagination framework. Leveraging large vision-language models in a dual-branch architecture, our approach achieves improved navigation performance with greater parameter efficiency, demonstrating potential applicability in embodied AI domains such as autonomous robotics, human-robot interaction, and assistive systems. 
Notwithstanding these advancements, several ethical and safety considerations remain paramount. Our experiments are conducted exclusively within controlled simulated environments to mitigate risks associated with unpredictable agent behavior. The propensity of large vision-language models to hallucinate or misinterpret environmental semantics highlights the necessity for rigorous evaluation and validation before real-world deployment. Additionally, as our method builds upon pretrained language models, it inherits challenges related to bias, fairness, and transparency, necessitating continuous efforts to identify and mitigate these risks. 

\clearpage
\medskip
{
    \small
    \bibliographystyle{plain}
    \bibliography{neurips_2025}
}

\end{document}